\pdfoutput=1

\documentclass[11pt]{article}

\usepackage[]{emnlp2021}

\usepackage{times}
\usepackage{latexsym}

\usepackage[T1]{fontenc}

\usepackage[utf8]{inputenc}

\usepackage{microtype}

\usepackage{amsmath}
\usepackage{amssymb}
\usepackage{graphicx}
\usepackage{makecell}
\usepackage{microtype}
\usepackage{multirow}
\usepackage{caption}
\usepackage{float}
\newcommand{\BigO}[1]{\ensuremath{\operatorname{O}\bigl(#1\bigr)}}

\title{Rethinking the Objectives of Extractive Question Answering}

\author{Martin Fajcik, Josef Jon,  Pavel Smrz\\
  Brno University of Technology\\
  {\tt \{ifajcik,ijon,smrz\}@fit.vutbr.cz} }

\date{}

\begin{document}
\maketitle
\begin{abstract}
    This work demonstrates that using the objective with independence assumption for modelling the span probability $P(a_s,a_e) = P(a_s)P(a_e)$ of span starting at position $a_s$ and ending at position $a_e$ has adverse effects. Therefore we propose multiple approaches to modelling joint probability $P(a_s,a_e)$ directly. Among those, we propose a compound objective, composed from the joint probability while still keeping the objective with independence assumption as an auxiliary objective. We find that the compound objective is consistently superior or equal to other assumptions in exact match. Additionally, we identified common errors caused by the assumption of independence and manually checked the counterpart predictions, demonstrating the impact of the compound objective on the real examples. Our findings are supported via experiments with three extractive QA models (BIDAF, BERT, ALBERT) over six datasets and our code, individual results and manual analysis are available online\footnote{\url{https://github.com/KNOT-FIT-BUT/JointSpanExtraction}.}.
\end{abstract}

\section{Introduction}

The goal of extractive question answering (EQA) is to find the span boundaries -- the start and the end of the span from text evidence, which answers a given question. Therefore, a natural choice of the objective to this problem is to model the probabilities of the span boundaries. 
In the last years, there was a lot of effort put into building better neural models underlying the desired probability distributions. However, there has been a little progress seen towards the change of the objective itself. 
For instance, the ``default'' choice of objective for modelling the probability over spans in SQuADv1.1~\cite{rajpurkar2016squad} -- maximization of independent span boundary probabilities $P(a_{s})P(a_{e})$ for answer at position $\langle a_{s}$,$a_{e} \rangle$ -- has stayed the same over the course of years in many influential works~\cite{xiong2016dynamic,seo2016bidirectional,chen2017reading,yu2018qanet, devlin2019bert, cheng2020probabilistic} since the earliest work on this dataset -- the submission of~\citet{wang2016machine}. Based on the myths of worse performance of different objectives, these works adopt the deeply rooted assumption of independence. However, this assumption may lead to obviously wrong predictions, as shown in Figure~\ref{fig:indep_error}. 
\begin{figure}[h!]
\includegraphics[width=0.48\textwidth, angle=0]{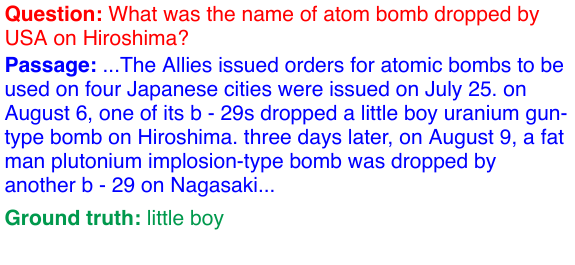}
\end{figure}

\begin{table}[h!]
\vspace{-6.5ex}
    \begin{center}
        \scalebox{0.78}{
            \begin{tabular}{cp{8cm}}
                \textbf{P} & \textbf{Predictions from BERT-base}                                                                                 \\ \hline
                33.3                & little boy uranium gun-type bomb on Hiroshima. three days later, on August 9, a fat man \\
                32.15                & little boy                 \\
                23.51                & fat man                                                                                              \\
                3.60               & a fat man                                                                              \\
                2.08                 & a little boy uranium gun - type bomb on hiroshima. three days later, on august 9, a fat man                                                                                              \\
                1.03                 & a little boy                                                                                              
                \end{tabular}
        }
    \captionof{figure}{\label{fig:indep_error} An example of an error which comes with an independence assumption. The model assigns high probability mass to boundaries around ``little boy'', and ``fat man'' answers. However, during decoding, the start of one and the end of another answer is picked up.}
    \end{center}
\end{table}
In addition, this assumption leads to degenerate distribution $P(a_{s}, a_{e})$, as high probability mass is assigned to many trivially wrong\footnote{We define 'trivially wrong' as not resembling any string form human would answer, e.g., the first or the second last answer of Figure \ref{fig:indep_error}.} answers. 

Some of the earlier work \cite{wang2016machine,weissenborn2017making} and recent approaches including large language representation models (LRMs) like XLNet \cite{yang2019xlnet}, ALBERT \cite{lan2019albert} or ELECTRA \cite{clark2020electra} started modelling the span probability via conditional probability factorization $P(a_e|a_s)P(a_s)$. However, is it unknown whether this objective improves any performance at all, as almost none of the recent works reported results on its effect, not even described its existence (except ELECTRA paper). Additionally, this objective requires beam search which slows down inference in test time. 
Exceptionally, \citet{lee2016learning} proposed one way for modelling $P(a_s, a_e)$ directly, but the approach was only sparsely adopted \cite{lee2019latent, khattab2020relevance}. This may be caused by the belief, that enumerating all possible spans has a large complexity \cite{cheng2020posterior}. However, in practice we find the complexity to be often similar to assumption of independence, when implementing the objective efficiently. We continue the in-depth discussion on complexity in Appendix \ref{sec:adressing_the_complexity}.

In this work, we try to break the myths about the objectives that have been widely used previously. We experiment with joint objective and we also introduce a new \textit{compound} objective, that deals with modelling joint probability $P(a_{s}, a_{e})$ directly while keeping the traditional independent objective as an auxiliary objective. We experiment with 5 different joint probability function realisations and find that with current LRMs, simple dot product works the best. However, we show that this is not a rule, and for some models, other function realisations might be better. The conducted experiments demonstrate that using compound objective is superior to previously used objectives across the various choices of models or datasets.

In summary, our work contributions are:
\begin{itemize}
    \item introduction of the compound objective and its comparison with the traditional objectives based on assumption of independence, conditional probability factorization, or direct joint probability,
    \item a thorough evaluation on the wide spectrum of models and datasets comparing different objectives supported by statistical tests,
    \item a manual analysis which provides closer look on the different impacts of independent and compound objectives.
\end{itemize}

\section{Probabilistic Assumptions for the Answer Span}
This section describes the common approach to the EQA, with its independent modelling of the answer span start and end positions. Secondly, it defines an assumption based on conditional factorization of span probability. Finally a function family for computing joint span probability and a combination of independent and joint assumption we call the \textit{compound} objective are proposed.  

The EQA can be defined as follows: Given a question $q$ and a passage or a set of passages $D$, find a string $a$ from $D$ such that $a$ answers the question $q$. This can be expressed by modelling a categorical probability mass function (PMF) that has its maximum in the answer start and end indices $a=\langle a_{s}, a_{e} \rangle$ from the passage $D$ as $P(a_{s}, a_{e}|q,D)$ for each question-passage-answer triplet $(q,D,a)$ from dataset $\mathcal{D}$. The parameters $\theta$ of such model can be estimated by minimizing maximum likelihood objective 
\begin{equation}
\label{eq:mle}
    - \sum_{(q,D,a)\in \mathcal{D}} \log P_{\theta}(a_{s}, a_{e}|q,D).
\end{equation}

During inference, the most probable answer span $\langle a_{s}, a_{e} \rangle$ is predicted.  Although there are works that were able to model the joint probability explicitly~\cite{lee2016learning}, modelling it directly results in a number of categories quadratic to the passage's length. Optimizing such models may be seen challenging, as there are often more classes than the amount of data points within the current datasets. Therefore, state-of-the-art approaches resort to independence assumption $P(a_{s}, a_{e}|q,D) = P_{\theta}(a_{s}|q,D)P_{\theta}(a_{e}|q,D)$. The factorized PMFs are usually computed by the model with shared parameters $\theta$, as introduced in~\citet{wang2016machine}. For most of the systems modelling the \textit{independent objective} with neural networks, the final endpoint probabilities\footnote{For brevity, $q,D$ dependencies are further omitted and bias terms are broadcasted along dimension $L$.} are derived from start/end position passage representations computed via shared model $\boldsymbol{H}_{s}$,$ \boldsymbol{H}_{e} \in \mathbb{R}^{d \times L}$ as shown for $b \in \{s,e\}$. 
\begin{equation}
    \label{eq:start_end_probs}
    P_{\theta}(a_{b}) = \mathrm{softmax}( \boldsymbol{w}_b^\top\boldsymbol{H}_{b} + \boldsymbol{b}_b)
\end{equation}
The passage representations $\boldsymbol{H}_{s}$,$ \boldsymbol{H}_{e}$ are often pre-softmax layer representations from neural network with passage and question at the input. Symbols $d$ and $L$ denote the model-specific dimension and the passage length, respectively.

Occasionally, the conditional factorization $P(a_{s}, a_{e}|q,D) = P_{\theta}(a_{s})P_{\theta}(a_{e}|a_{s})$ is considered instead. The probabilities of span's start and end are computed the same way as in equation \ref{eq:start_end_probs}. The difference is in the end representations $\boldsymbol{H}_{e} = f(a_{s})$, which now must be the a function of span's start $a_{s}$.

\subsection{Joint Assumptions}
However, one does not need to apply simplifying assumptions and instead compute joint probability directly. We define a family of \textit{joint} probability functions $P_{\theta}(a_{s},a_{e})$ with an arbitrary vector-to-vector similarity function $f_{sim}$ used for obtaining each span score (e.\,g., the dot product $\boldsymbol{H}_{s}^\top \boldsymbol{H}_{e}$)\footnote{Here, we slightly abuse the notation for the sake of generality. See Subsection \ref{section:applied_models} for specific applications.}.
\begin{equation}
    \label{eq:joint}
    P_{\theta}(a_{s},a_{e}) = \mathrm{softmax}(\mathrm{vec}(f_{sim}(\boldsymbol{H}_{s},\boldsymbol{H}_{e})))
\end{equation}

Finally, we define a multi-task \textit{compound} objective  \eqref{eq:joint_indep_loss} composing the joint and independent probability formulations,  computed via a shared model $\theta$.

\begin{equation}
\label{eq:joint_indep_loss}
    -\sum_{(q,D,a)\in \mathcal{D}} \log P_{\theta}(a_{s}, a_{e}) P_{\theta}(a_{s})P_{\theta}(a_{e}) 
\end{equation}
Here $P(a_{s})P(a_{e})$ can be seen as a auxiliary objective for the more complex joint objective $P_{\theta}(a_{s}, a_{e})$ used for decoding in test time. Empirically, we found the \textit{compound} objective to be superior or equal to other assumptions.

\section{Experimental Setup}
We use Transformers \cite{wolf2019huggingface} for language representation model (LRM) implementation. Our experiments were done on 16GB GPUs using PyTorch~\cite{paszke2019pytorch}. For experiments with LRMs, we used Adam optimizer with a decoupled weight decay~\cite{loshchilov2018decoupled}. The used hyperparameters were the same as the SQuADv1.1 default hyperparameters as proposed by specific LRM authors through all our datasets. For BIDAF, we tuned hyperparameters using Hyperopt \cite{bergstra2013hyperopt} separately for independent and compound objectives\footnote{We used $f_{madd}$ similarity during parameter tuning.}.
See Appendix \ref{section:hyperparameters} for further details.

In all our experiments, we apply \textit{length filtering} (LF). Therefore, probabilities $P(a_{s} = i, a_{e} =j)$ are set to $0$ iff $j-i>\zeta$, where $\zeta$ is a length threshold. Following~\citet{devlin2019bert}, we set $\zeta = 30$ in all of our experiments.

\subsection{Similarity Functions}
\label{section:exploration_of_similarity_functions}
Here we sum up the definitions of similarity functions presented in the paper. We experimented with 5 similarity functions. For each start representation $\boldsymbol{h_s} \in \mathbb{R}^d$ and end representation $\boldsymbol{h_e} \in \mathbb{R}^d$, both column vectors from the matrix of boundary vectors $\boldsymbol{H}_s $, $\boldsymbol{H}_e \in \mathbb{R}^{d \times L}$ respectively. Note that $d$ here is model specific dimension, $L$ is passage length, $\circ$ denotes elementwise multiplication and $;$ denotes concatenation. The similarity functions above these representations are defined as:
\begin{itemize}
    \item A dot product:
    \begin{equation}
        f_{dot}(\boldsymbol{h_s},\boldsymbol{h_e}) = \boldsymbol{h_s}^{\top}\boldsymbol{h_e}
    \end{equation}
    \item A weighted dot product:
    \begin{equation}
        f_{wdot}(\boldsymbol{h_s},\boldsymbol{h_e}) = \boldsymbol{w}^{\top}[\boldsymbol{h_s } \circ \boldsymbol{h_e}]
    \end{equation}
    \item An additive similarity:
    \begin{equation}
        f_{add}(\boldsymbol{h_s},\boldsymbol{h_e}) = \boldsymbol{w}^{\top}[\boldsymbol{h_s};\boldsymbol{h_e}]
    \end{equation}
    \item An additive similarity combined with weighted product:
    \begin{equation}
        f_{madd}(\boldsymbol{h_s},\boldsymbol{h_e}) = \boldsymbol{w}^{\top}[\boldsymbol{h_s};\boldsymbol{h_e};\boldsymbol{h_s } \circ \boldsymbol{h_e}]
    \end{equation}
    \item A multi-layer perceptron (MLP) as proposed by \citet{lee2019latent}:
    \begin{equation}
        f_{MLP}(\boldsymbol{h_s},\boldsymbol{h_e}) = \boldsymbol{w}^{\top}\sigma(\boldsymbol{W}[\boldsymbol{h_s};\boldsymbol{h_e}]+\boldsymbol{b})+\boldsymbol{b}_2
    \end{equation}
    where $\sigma(x) = ln(relu(x))$ and $ln$ denotes layer normalization \cite{ba2016layer}.
\end{itemize}

\subsection{Applied Models}
\label{section:applied_models}
Our experiments are based on three EQA models:

\noindent\textbf{BERT}-base~\cite{devlin2019bert} and \textbf{ALBERT}-xxlarge~\cite{lan2019albert} are LRMs based on the self-supervised pretraining objective. During fine-tuning, each model receives the concatenation of question and passage are given as input. Outputs $\boldsymbol{H} \in \mathbb{R}^{d \times L}$ corresponding to the passage inputs of length $L$ are then reduced to boundary probabilities by two vectors $\boldsymbol{w_s}$, $\boldsymbol{w_e}$ as
$P(a_b) = \mathrm{softmax}(\boldsymbol{w}_b^{\top} \boldsymbol{H}  + \boldsymbol{b}_b)$ where $b \in \{s,e\}$.
To compute  joint probability $P(a_{s}, a_{e})$, start representations are computed using $\boldsymbol{W} \in \mathbb{R}^{d \times d}$ and $\boldsymbol{b} \in \mathbb{R}^{d}$ (broadcasted) as $\boldsymbol{H}_s=\boldsymbol{W}\boldsymbol{H} + \boldsymbol{b}$ and end representations as $\boldsymbol{H}_e=\boldsymbol{H}$. A dot product $f_{dot}$ is used as the similarity measure.
\begin{equation}
    P(a_{s}, a_{e}) = \mathrm{softmax}(\mathrm{vec}(\boldsymbol{H}_s^{\top}\boldsymbol{H}_e)) 
\end{equation}
For modelling conditional probability factorization objective, we follow the implementation from \cite{lan2019albert}, and provide exact details in the Appendix \ref{appendix:conditional_obj}.

\noindent \textbf{BIDAF}~\cite{seo2016bidirectional} dominated the state-of-the-art systems in 2016 and motivated a lot of following research work~\cite{clark2017simple,yu2018qanet}. Question and passage inputs are represented via the fusion of word-level embeddings from GloVe~\cite{pennington2014glove} and character-level word embeddings obtained via a convolutional neural network. Next, a recurrent layer is applied to both. Independently represented questions and passages are then combined into a common representation via two directions of attention over their similarity matrix~$\boldsymbol{S}$. The similarity matrix is computed via multiplicative-additive interaction~\eqref{eq:multi_add_sim} between each pair of question vector~$\boldsymbol{q_i}$  and passage vector~$\boldsymbol{p_j}$, where $;$ denotes concatenation and $\circ$ stands for the Hadamard product.
\begin{equation}
\label{eq:multi_add_sim}
    \boldsymbol{S_{ij}}=f_{madd}(\boldsymbol{q_i},\boldsymbol{p_j}) = \boldsymbol{w}^{\top}[\boldsymbol{q_i};\boldsymbol{p_j};\boldsymbol{q_i } \circ \boldsymbol{p_j}]
\end{equation}
Common representations are then concatenated together with document representations yielding $\boldsymbol{G}$ and passed towards two more recurrent layers producing $\boldsymbol{M}$ and ${\boldsymbol{M}}^2$ -- first to obtain answer-start representations $\boldsymbol{H}_s = [\boldsymbol{G};\boldsymbol{M}]$ and second to obtain answer-end representations\footnote{For details, see formulae~2 to~4 in \citet{seo2016bidirectional}.} $\boldsymbol{H}_e = [\boldsymbol{G};\boldsymbol{M}^2]$.
The joint probability $P(a_{s}, a_{e})$ is then computed over scores from vectorized similarity matrix of $\boldsymbol{H}_s$ and $\boldsymbol{H}_e$ using the 2-layer feed-forward network $f_{MLP}$ as a similarity function.

\subsection{Datasets}
\begin{table}[ht!]
\begin{tabular}{c|cc}
\textbf{Dataset}  & \multicolumn{1}{c|}{\textbf{Train}} & \textbf{Test} \\ \hline
SQuADv1.1         & 87,599                              & 10,570         \\
SQuADv2.0         & 130,319                             & 11,873         \\
Adversarial SQuAD & -                                   & 3,560          \\
Natural Questions & 104,071                             & 12,836        \\
NewsQA            & 74,160                              & 4,212         \\
TriviaQA          & 61,688                              & 7,785       
\end{tabular}
\caption{\label{tab:dataset_statistics}Number of examples per each dataset used in this paper.}
\end{table}
We evaluate our approaches on a wide spectrum of datasets. We do not split development datasets, as we use fixed hyperparameters with fixed amount of steps and use last checkpoint for our LRM experiments. This also makes our results directly comparable to other works \cite{devlin2019bert, lan2019albert}. The statistics to all datasets are shown in Table \ref{tab:dataset_statistics}.
To focus only on the extractive part of QA and to keep the format the same, we use curated versions of the last 3 datasets as released in MrQA shared task~\cite{fisch2019mrqa}.

\textbf{SQuADv1.1}~\cite{rajpurkar2016squad} is a popular dataset composed from question, paragraphs and answer span annotation collected from the subset of Wikipedia passages.

\textbf{SQuADv2.0}~\cite{rajpurkar2018know} is an extension of SQuADv1.1 with additional 50k questions and passages, which are topically similar to the question, but do not contain an answer.

\textbf{Adversarial SQuAD}~\cite{jia2017adversarial} tests, whether the system can answer questions about paragraphs that contain adversarially inserted sentences, which are automatically generated to distract computer systems without changing the correct answer or misleading humans. In particular, our system is evaluated in \textsc{AddSent} adversary setting, which runs the model as a black box for each question on several paragraphs containing different adversarial sentences and picks the worst answer.

\textbf{Natural Questions}~\cite{kwiatkowski2019natural} dataset consists of real users queries obtained from Google search engine. Each example is accompanied by a relevant Wikipedia article found by the search engine, and human annotation for long/short answer. The long answer is typically the most relevant paragraph from the article, while short answer consists of one or multiple entities or short text spans. We only consider short answers in this work.

\textbf{NewsQA}~\cite{trischler2017newsqa} is a crowd-sourced dataset based on CNN news articles. Answers are short text spans and the questions are designed such that they require reasoning and inference besides simple text matching.

\textbf{TriviaQA}~\cite{joshi2017triviaqa} consists of question-answer pairs from 14 different trivia quiz websites and independent evidence passages collected using Bing search from various sources such as news, encyclopedias, blog posts and others. Additional evidence is obtained from Wikipedia through entity linker.

\subsection{Statistical Testing}
To improve the soundness of the presented results, we use several statistical tests. An exact match (EM) metric can be viewed as an average of samples from Bernoulli distribution. As stated via central limit theorem, a good assumption might be the EM comes from the normal distribution. We train 10 models for each presented LRM's result, obtaining 10 EMs for each sample. \textit{Anderson-Darling} normality test \cite{stephens1974edf} is used to check this assumption -- whether the sample truly comes from the normal distribution. Then we use the \textit{one-tailed paired t-test} to check whether the case of improvement is significant. The improvement is significant iff $\textrm{p-value} <0.05$. We use the reference implementation from \citet{dror2018hitchhiker}.

\section{Results and Discussion}
\label{section:results}
\begin{table*}[ht!]
    \begin{center}
        \scalebox{0.94}{
            \begin{tabular}{|c|c|c|c|c|c|c|c|}
                \hline
                \textbf{Model} & \textbf{Obj} & \textbf{SQ1}                  & \textbf{SQ2}                  & \textbf{AdvSQ}                & \textbf{TriviaQA}    & \textbf{NQ}         & \textbf{NewsQA} \\ \hline
                    \multirow{4}{*}{BERT}     & I            & 81.31/88.65                   & \textbf{73.89}/76.74          & 47.04/52.62                   & 62.88/69.85          & 65.66/78.20         & 52.39/67.17     \\ \cline{2-8} 
                               & J            & 81.33/88.13                   & 72.66/75.04                   & 48.10/53.54                    &\textbf{ 63.93}/69.90          & \textbf{67.75}/78.70         & 52.73/66.41   \\ \cline{2-8} 
                               & JC           & 81.22/88.29                  & 71.51/74.38                             & 46.07/51.35                   & 62.82/69.94          & 66.48/77.34         & 52.39/67.05     \\ \cline{2-8} 
                               & I+J          & \textit{\textbf{81.83}}/88.52          & 73.53/76.14                   & \textit{\textbf{48.32}}/53.47          & \textit{63.73}/69.75& \textit{\textbf{67.75}}/78.81         & \textbf{\textit{52.96}}/66.83     \\ \hline \hline
                    \multirow{4}{*}{ALBERT}   & I            & 88.55/94.62                   & 87.07/90.02                   & 68.12/73.54                   & 74.7/80.33          & 70.78/83.42         & 59.95/75.0     \\ \cline{2-8} 
                               & J            & 88.84/94.64                   & 86.87/89.71                   & 68.90/74.17                   & 75.11/80.41                    & \textbf{73.36/84.01 }                  & 60.19/74.28               \\ \cline{2-8} 
                               & JC         & 88.60/94.59                   & 86.78/89.73                             & 68.0/73.25                  & -                    & 72.33/83.35                  & 58.52/72.74               \\ \cline{2-8} 
                               & I+J          & \textit{\textbf{89.02}}/94.77          & \textbf{87.13}/89.98          & \textit{\textbf{69.57}}/74.76                   & \textit{\textbf{75.31}}/80.43 & \textit{73.32}/84.08& \textbf{ \textit{60.41}}/74.46\\ \hline
            \end{tabular}
        }
    \end{center}
    \caption{\label{tab:main_results} EM/F1 results of different objectives through the spectrum of datasets. Bold results mark best EM across the objectives. Italicised I+J results mark significant improvement over the independent objective. }
\end{table*}

We now show the effectiveness of proposed approaches. Each of the presented results is averaged from 10 training runs.

\begin{table}[H]
\centering
\begin{tabular}{|c|c|c|c|c|}
\hline
 & \textbf{EM} & \textbf{F1} & \textbf{EM} & \textbf{F1} \\ \hline
 I & 66.16 & 76.19 & 81.31 & 88.65 \\ \hline\hline
 I+J& \multicolumn{2}{c|}{BIDAF} & \multicolumn{2}{c|}{BERT} \\ \hline
$f_{dot}$ & 64.30 & 73.84 & \textbf{81.83} & \textbf{88.52} \\ \hline
$f_{add}$ & 66.04 & 75.10 & 81.52 & 88.47 \\ \hline
$f_{wdot}$ & 66.10 & 75.16 & 81.35 & 88.29 \\ \hline
$f_{madd}$ & 66.11 & 75.23 & 81.45 & 88.44 \\ \hline
$f_{MLP}$ & \textbf{66.96} & \textbf{75.90} & 81.61 & 88.44\\ \hline
\end{tabular}
\caption{A comparison of similarity functions in the models trained via \textit{compound} objective (I+J) and \textit{independent} objective (I).}
\label{tab:sim_functions}
\end{table}
\textbf{Similarity functions}. We analyzed an effect of different similarity functions over all models in Table \ref{tab:sim_functions}. We found different similarity functions to work better with different architectures. Namely, for BIDAF, most of similarity functions work equally or worse than independent objective. Exceptionally, $f_{MLP}$ works significantly better. This is surprising especially because we tuned the hyperparameters with the $f_{madd}$ function. For BERT, most of the similarity functions performed better than the independent objective and simple dot-product $f_{dot}$ improved significantly better above all. We choose $f_{MLP}$ for BIDAF and $f_{{dot}}$ for our LRMs for the rest of experiments.

\textbf{Comparison of objectives}. Our main results -- the performance of \textit{independent} (I), \textit{joint} (J), \textit{joint-conditional} (JC) and \textit{compound} (I+J) objectives -- are shown in Table~\ref{tab:main_results}.
We note the largest improvements can be seen for an exact match (EM) performance metric. In fact, in some cases objectives modelling joint PMF lead to degradation of F1, while improving EM (e.g., on SQuADv1.1 and NewsQA datasets for BERT). Upon manual analysis of BERT's predictions based on 200 differences between independent and compound models on SQuADv1.1, we found that in 10 cases (5\%) the independent model chooses larger span encompassing multiple potential answers, thus obtaining non-zero F1 score. In 9 out of 10 of these cases, we found the compound model to pick just one of these potential answers\footnote{For instance, in Table \ref{tab:examples_of_errors}, row 4, column 3, we consider 2,000; 40,000; 2,200; 1,294 and 427 as potential answers.}, obtaining either full match or no F1 score at all. We found no cases of compound model encompassing multiple potential answers in analyzed sample.

Next, we remark that compound objective outperformed others in most of our experiments. In BERT case, the compound objective performed significantly better than independent objective on 5 out of 6 datasets. In ALBERT case, the compound objective performed significantly better than independent objective 5 from 6 times and it was on par in the last case.
Comparing compound to joint objective in BERT case, the two behave almost equally, with compound objective significantly outperforming joint objective on the two SQuAD datasets and no significant differences for the other 4 datasets.
However, in ALBERT case, the compound objective significantly improves results over joint objective in all but one case and is on par in this last case. 
\begin{table*}[!ht]
    \centering
    \begin{tabular}{|c|c|c|c|c|c|}
\hline
\textbf{Model} &    & \textbf{I}                    & \textbf{J}  & \textbf{JC} & \textbf{I+J} \\ \hline
\multirow{2}{*}{BIDAF}               
               & -  & 65.85/75.94                   & -           & -           & 66.95/75.89  \\ \cline{2-6} 
               & LF & \textbf{66.16}/\textbf{76.19 }& 58.24/67.42 & -           & 66.96/75.90  \\ \hline\hline
\multirow{2}{*}{BERT}
               & -  & 80.98/88.40                   & 81.30/88.11 & 81.16/88.25 & 81.80/88.50  \\ \cline{2-6} 
               & LF & \textbf{81.31}/\textbf{88.65} & 81.33/88.13 & 81.22/88.29 & 81.83/88.52  \\ \hline\hline
\multirow{2}{*}{ALBERT}
               & -  & 88.39/94.51                   & 88.82/94.64 & 88.57/94.57 & 89.01/94.77  \\ \cline{2-6} 
               & LF & \textbf{88.55}/\textbf{94.63} & 88.84/94.64 & 88.60/94.59 & 89.02/94.77  \\ \hline
\end{tabular}
    \caption{\label{tab:results_of_filtering} SQuADv1.1 EM/F1 results with length filtering (LF) computed from the same set of checkpoints. Differences larger than 0.1 are in bold.}
\end{table*}

\textbf{Conditional objective}. 
Our implementation of the conditional objective performs even or worse than independent objective in most cases. Upon investigation we found the model tends to be overconfident about start predictions and underconfident about its end predictions, often assigning high probability to single answer-start. In Table \ref{tab:co_analysis}, we analyze the top-5 most probable samples from BERT on each example of SQuADv1.1 dev data. We found that on average the conditional model kept it top-1 start prediction in 90\% of subsequent top-2 to top-5 less probable answers, but kept its top-1 end prediction only in 4\% of top-2 to top-5 subsequent answers. We found this statistic to be on par for start/end prediction for different objectives. Interestingly, the table also reveals that independent objective contains less diverse start/end tokens than joint objectives.
\begin{table}[!ht]
\centering
\begin{tabular}{|c|c|c|l}
\cline{1-3}
\textbf{Model} & \textbf{Start Token} & \textbf{End Token} &  \\ \cline{1-3}
I              & 43.76\%        & 45.66\%      &  \\ \cline{1-3}
JC             & 90.12\%        & 3.9\%        &  \\ \cline{1-3}
J              & 33.71\%        & 35.91\%      &  \\ \cline{1-3}
I+J            & 34.95\%        & 37.27\%      &  \\ \cline{1-3}
\end{tabular}
    \caption{\label{tab:co_analysis} Proportion of samples, on which top-1 prediction start/end token was kept as start/end token also in top-2 to top-5 subsequent predictions.}
\end{table}

\textbf{Large improvements and degradation}. 
Upon closer inspection of results, we found possible reasons for result degradation of the compound model on SQuADv2.0, and also its large improvements gained on NQ dataset. 

For SQuADv2.0, the accuracies of no-answer detection for independent/joint/compound objectives in case of BERT models are 79.89/78.12/79.32. We found the same trend for ALBERT. We hypothesize, that this inferior performance of joint and compound models may be caused by the model having to learn a more complex problem of $K^2$ classes of all possible spans over input document, which is often more (e.g. for $K=512$) than the size of the datasets, leaving the less of ``model capacity'' to this another task. To confirm that compound model is better at answer extraction step, we run all 10 checkpoints trained on SQuADv2.0 data with an answer, while masking model's no-answer option. The results shown in Table \ref{tab:squad2_models_on_answer_perf} support this hypothesis.

\begin{table}
    \centering
    \begin{tabular}{|c|c|c|c|}
        \hline
        \textbf{\textbf{}}      & \textbf{Objective} & \textbf{EM}    & \textbf{F1}    \\ \hline
        \multirow{3}{*}{BERT}   & I         & 80.70 & 88.71 \\ \cline{2-4} 
                                & J         & 81.38 & 81.51 \\ \cline{2-4} 
                                & I+J       & 81.51 & 88.69 \\ \hline \hline
        \multirow{3}{*}{ALBERT} & I         & 87.40 & 94.10 \\ \cline{2-4} 
                                & J         & 87.74 & 94.31 \\ \cline{2-4} 
                                & I+J       & 87.90 & 94.38 \\ \hline
    \end{tabular}
    \caption{\label{tab:squad2_models_on_answer_perf} Performance of SQuADv2.0 models on answerable examples of SQuADv2.0.}
\end{table}

On the other side, we found the large improvements over NQ might be exaggerated by the evaluation approach of MRQA, wherein the case of multi-span answers, choosing one of the spans from multi-span answer counts as correct. Upon closer result inspection, we found that the independent model here was prone to select the start of one span from multi-span answer and end of different span from multi-span answer. To quantify this behavior, we annotated 100 random predictions with multi-span answers in original NaturalQuestions on whether they pick just one span from multi-span answer (which follows from the MRQA formulation) or they encompass multiple spans. For independent/compound objectives we found 59/77 cases of picking just one of the spans and 22/4 cases of encompassing multiple spans from multi-span answer for BERT model and 57/81 and 33/10 cases for ALBERT respectively.

\textbf{Length filtering heuristic}. Additionally, we found the benefit from the commonly used length filtering (LF) heuristic is negligible for models trained via any joint objective, as shown in Table \ref{tab:results_of_filtering}. Therefore, we find it unnecessary to use the heuristic anymore. In this experiment, we also include our results with BIDAF, which show significant improvement of compound objective on SQuADv1.1 dataset from other approaches.

\begin{table*}[!ht]
    \begin{center}
        \scalebox{0.66}{
            \begin{tabular}{|p{4cm}|p{10cm}|p{4cm}|p{2cm}|p{2cm}|}
                \hline
                \multicolumn{1}{|c|}{\textbf{Question}} & \multicolumn{1}{c|}{\textbf{Passage}} & \multicolumn{1}{p{4cm}|}{\textbf{Independent}} & \multicolumn{1}{p{2cm}|}{\textbf{Compound}} & \multicolumn{1}{p{2cm}|}{\textbf{Ground Truth}} \\ 
                \hline
                What company won a free advertisement due to the QuickBooks contest? & QuickBooks sponsored a "Small Business Big Game" contest, in which Death Wish Coffee had a 30-second commercial aired free of charge courtesy of QuickBooks. Death Wish Coffee beat out nine other contenders from across the United States for the free advertisement. &
                Death Wish Coffee had a 30-second commercial aired free of charge courtesy of QuickBooks. Death Wish Coffee &
                Death Wish Coffee &
                Death Wish Coffee \\ 
                \hline
                In what city's Marriott did the Panthers stay? &
                The Panthers used the San Jose State practice facility and stayed at the San Jose Marriott. The Broncos practiced at Stanford University and stayed at the Santa Clara Marriott. &
                San Jose State practice facility and stayed at the San Jose & San Jose &
                San Jose \\ 
                \hline
                What was the first point of the Reformation? &
                Luther's rediscovery of "Christ and His salvation" was the first of two points that became the foundation for the Reformation. His railing against the sale of indulgences was based on it. &
                Christ and His salvation" &
                Christ and His salvation &
                Christ and His salvation \\ 
                \hline
                How many species of bird and mammals are there in the Amazon region? &
                The region is home to about 2.5 million insect species, tens of thousands of plants, and some 2,000 birds and mammals. To date, at least 40,000 plant species, 2,200 fishes, 1,294 birds, 427 mammals, 428 amphibians, and 378 reptiles have been scientifically classified in the region. One in five of all the bird species in theworld live in the rainforests of the Amazon, and one in five of the fishspecies live in Amazonian rivers and streams. Scientists have describedbetween 96,660 and 128,843 invertebrate species in Brazil alone. &
                2,000 birds and mammals. To date, at least 40,000 plant species, 2,200 fishes, 1,294 birds, 427 &
                427 &
                2,000 \\ 
                \hline
                What was found to be at fault for the fire in the cabin on Apollo 1 regarding the CM design? &
                NASA immediately convened an accident review board, overseen by both houses of Congress. While the determination of responsibility for the accident was complex, the review board concluded that "deficiencies existed in Command Module design, workmanship and quality control." At the insistence of NASA Administrator Webb, North American removed Harrison Storms as Command Module program manager. Webb also reassigned Apollo Spacecraft Program Office (ASPO) Manager Joseph Francis Shea, replacing him with George Low. &
                deficiencies existed in Command Module design, workmanship and quality control." &
                Harrison Storms &
                deficiencies \\ 
                \hline
            \end{tabular}
        }
    \end{center}
    \caption{\label{tab:examples_of_errors} Examples of predictions from SQuADv1.1 using BERT trained with independent and compound objective.}
\end{table*}

\section{Analysis}

\label{section:analysis}
Apart from example in Figure~\ref{fig:indep_error}, we provide more examples of different predictions\footnote{We chose to analyze the different predictions, as the model is usually more uncertain in these borderline cases.} between models trained with independent and compound objective in Table~\ref{tab:examples_of_errors}. In general, by doing manual analysis of errors, we noticed three types of trivially wrong errors being fixed by the compound objective model in BERT:
\begin{enumerate}
  \item Uncertainty of the model causes it to assign high probabilities to two different answer boundaries. During decoding the start/end boundaries of two different answers are picked up (fourth row in Table~\ref{tab:examples_of_errors}).
  \item The model assigns high probability to answer surrounded by the paired punctuation marks (e.\,g. quotes). It chooses the answer without respecting the symmetry between paired punctuation marks (third row of Table~\ref{tab:examples_of_errors}).
  \item Uncertainty of the model causes it to assign high probabilities to two spans containing the same answer string. This is the special case of problem (1) -- while the model often chooses the correct answer, the boundaries of two different spans are selected (first row of Table~\ref{tab:examples_of_errors}).
\end{enumerate}

To quantify an occurrence of these errors, we study our best BERT and ALBERT checkpoint predictions for SQuADv1.1 validation data. For BERT, we found the most frequently occurring is the error type (1), for which we manually annotated 200 random differences between independent and compound model predictions. We found \textit{5\%} of them to be the case of this error of the independent, and no case of this error for the compound model. Interestingly, 4 out of 10 of these cases were questions clearly asking about single entity, while independent model answered multiple entities, e.g., \textit{Q:Which male child of Ghengis Khan and Börte was born last? A:Chagatai (1187—1241), Ögedei (1189—1241), and Tolui}. For the error type (2), we filtered all prediction differences (more than 1300 for BERT and ALBERT) down to cases, where either independent or compound prediction contained non-alphanumeric paired punctuation marks, which resulted in less than 30 cases for each. For BERT, \textit{37\%} independent predictions from these cases contained an error type (2), while again no paired punctuation marks errors were observed for compound objective.

For the error type (3), we filter prediction differences down to cases, where independent or compound prediction contained the same prefix and suffix of length at least 2 (only 9 and 5 cases for BERT and ALBERT). From these, error type (3) occurred in 3 cases for BERT and in 1 case for ALBERT in case of independent and again we found no case for the compound for both models. Note the error type (3) can be fully alleviated by marginalizing over probabilities of top-K answer spans during the inference, as in \cite{das2019multi, cheng2020probabilistic} (see Appendix \ref{appendix:sff} for details).
Interestingly, for ALBERT, we found only negligible amount of errors of type (1) and (2) for both objectives\footnote{The full difference of BERT's and ALBERT's predictions and manual analysis can be found in the supplementary.}.

\begin{figure}[ht!]
    \centering
    \includegraphics[width=\columnwidth]{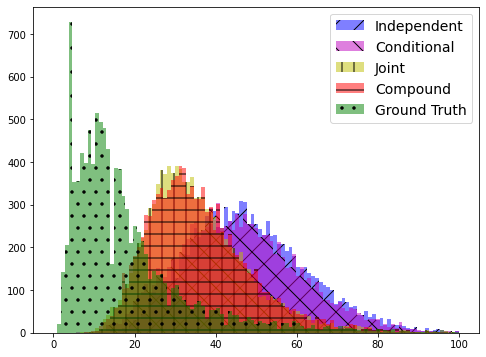}
    \caption{Histograms of average character length of top-20 predicted answers from BERT trained with different objectives compared with character length of ground-truth answers.
    \label{fig:indep_vs_joint_hist}}

\end{figure}

During manual analysis, we observed that, an uncertain models with an independent objective are prone to pick large answer spans. To illustrate, that spans retrieved with approaches modelling joint probability differ, we took the top 20 most probable spans from each model and averaged their length.

This was done for each example in the SQuADv1.1 test data. The histogram of these averages is shown in Figure~\ref{fig:indep_vs_joint_hist}. For a fair comparison, these predictions were filtered via length filtering.

\section{Related Work}
One of the earliest works in EQA from~\citet{wang2016machine} experimented with generative models based on index sequence generation via pointer networks~\cite{vinyals2015pointer} and now traditional boundary models that focus on the prediction of start/end of an answer span. Their work shown substantial improvement of conditional factorization boundary models over the index sequence generative models. 

Followup work on EQA \cite{seo2016bidirectional,chen2017reading, clark2017simple,yu2018qanet, devlin2019bert, cheng2020probabilistic} and others considered using the assumption of independence in their objectives.

\citet{xiong2016dynamic} explored an iterative boundary model. They used RNN and a highway maxout network to decode start/end of span independently in multiple timesteps, each time feeding the RNN with predictions from the previous time step until the prediction was not changing anymore. In their following work~\citet{xiong2017dcn+} combined their objective with a reinforcement learning approach, in which the decoded spans from each timestep were treated as a trajectory. They argued that cross-entropy is not reflecting F1 performance well enough, and defined a reward function equal to F1 score. Finally, they used policy gradients as their auxiliary objective, showing 1\% improvement in the terms of F1 score. 

Authors of recent LRMs like XLNet \cite{yang2019xlnet}, ALBERT \cite{lan2019albert} or ELECTRA \cite{clark2020electra} use conditional probability factorization $P(a_e|a_s)P(a_s)$ for answer extraction in some cases\footnote{For instance, ALBERT uses conditional objective for SQuADv2.0, but not for SQuADv1.1.}. Although the objective is not described in mentioned papers (except for ELECTRA), we follow the recipe for modelling the conditional probability from their implementation in this work. We believe this is the first official comparison of this objective w.r.t. others.

The most similar to our work is \textsc{RaSoR} system~\cite{lee2016learning}. In their work, authors compared various objectives -- binary answer classification of every input token, BIO sequence classification with CRF layer on top of their model, and most importantly joint objective, which turns out to work the best. However, in our experiments, training with the joint objective alone does not always perform that well. For BIDAF, we failed to find the hyperparameters for model to converge to results similar to different approaches.

\section{Conclusion}
The paper closely studies the objectives used within the extractive question answering (EQA).
It identifies commonly used independent probability model as a source of trivially wrong answers. As a remedy it experiments with various ways of learning the joint span probability. Finally it shows how the \textit{compound} objective -- the combination of independent and joint probability in objective -- improves statistical EQA systems across 6 datasets without using any additional data. 
Using the proposed approach, we were able to reach significant improvements through the wide spectrum of datasets, including +1.28 EM on Adversarial SQuAD and +2.07 EM on NaturalQuestions for BERT-base. We performed a thorough manual analysis to understand what happened to trivially wrong answers, and we found most of the cases disappear. We also found that independent models tend to ``overfit'' to F1 metric by encompassing multiple possible answer spans, which would explain the effect of joint objectives improving the EM far more significantly than the F1.  We shown the samples from joint model contain the greatest start/end token diversity. We further hypothesize that having diverse answers may be especially beneficial towards answer reranking step commonly used in QA \cite{fajcik2021r2d2, iyer-etal-2021-reconsider}.
In addition, we also identified the reason for performance decrease with \textit{compound} objective on SQuADv2.0 -- no-answer classifier trained within the same model performs worse -- and we leave the solution for this deficiency for future work. 

\section*{Acknowledgments}
This work was supported by the Czech Ministry of Education, Youth and Sports, subprogram INTERCOST, project code: LTC18054.
The computation used the infrastructure supported by the Ministry of Education, Youth and Sports of the Czech Republic through the e-INFRA CZ (ID:90140).

\bibliography{anthology,custom}

\begin{thebibliography}{36}
\expandafter\ifx\csname natexlab\endcsname\relax\def\natexlab#1{#1}\fi

\bibitem[{Ba et~al.(2016)Ba, Kiros, and Hinton}]{ba2016layer}
Jimmy~Lei Ba, Jamie~Ryan Kiros, and Geoffrey~E Hinton. 2016.
\newblock Layer normalization.
\newblock \emph{arXiv preprint arXiv:1607.06450}.

\bibitem[{Bergstra et~al.(2013)Bergstra, Yamins, Cox
  et~al.}]{bergstra2013hyperopt}
James Bergstra, Dan Yamins, David~D Cox, et~al. 2013.
\newblock Hyperopt: A python library for optimizing the hyperparameters of
  machine learning algorithms.
\newblock In \emph{Proceedings of the 12th Python in science conference},
  volume~13, page~20. Citeseer.

\bibitem[{Chen et~al.(2017)Chen, Fisch, Weston, and Bordes}]{chen2017reading}
Danqi Chen, Adam Fisch, Jason Weston, and Antoine Bordes. 2017.
\newblock \href {https://doi.org/10.18653/v1/P17-1171} {Reading {W}ikipedia to
  answer open-domain questions}.
\newblock In \emph{Proceedings of the 55th Annual Meeting of the Association
  for Computational Linguistics (Volume 1: Long Papers)}, pages 1870--1879,
  Vancouver, Canada. Association for Computational Linguistics.

\bibitem[{Cheng et~al.(2020)Cheng, Chang, Lee, and
  Toutanova}]{cheng2020probabilistic}
Hao Cheng, Ming-Wei Chang, Kenton Lee, and Kristina Toutanova. 2020.
\newblock \href {https://doi.org/10.18653/v1/2020.acl-main.501} {Probabilistic
  assumptions matter: Improved models for distantly-supervised document-level
  question answering}.
\newblock In \emph{Proceedings of the 58th Annual Meeting of the Association
  for Computational Linguistics}, pages 5657--5667, Online. Association for
  Computational Linguistics.

\bibitem[{Cheng et~al.(2021)Cheng, Liu, Pereira, Yu, and
  Gao}]{cheng2020posterior}
Hao Cheng, Xiaodong Liu, Lis Pereira, Yaoliang Yu, and Jianfeng Gao. 2021.
\newblock \href {https://www.aclweb.org/anthology/2021.naacl-main.85}
  {Posterior differential regularization with f-divergence for improving model
  robustness}.
\newblock In \emph{Proceedings of the 2021 Conference of the North American
  Chapter of the Association for Computational Linguistics: Human Language
  Technologies}, pages 1078--1089, Online. Association for Computational
  Linguistics.

\bibitem[{Clark and Gardner(2018)}]{clark2017simple}
Christopher Clark and Matt Gardner. 2018.
\newblock \href {https://doi.org/10.18653/v1/P18-1078} {Simple and effective
  multi-paragraph reading comprehension}.
\newblock In \emph{Proceedings of the 56th Annual Meeting of the Association
  for Computational Linguistics (Volume 1: Long Papers)}, pages 845--855,
  Melbourne, Australia. Association for Computational Linguistics.

\bibitem[{Clark et~al.(2020)Clark, Luong, Le, and Manning}]{clark2020electra}
Kevin Clark, Minh{-}Thang Luong, Quoc~V. Le, and Christopher~D. Manning. 2020.
\newblock \href {https://openreview.net/forum?id=r1xMH1BtvB} {{ELECTRA:}
  pre-training text encoders as discriminators rather than generators}.
\newblock In \emph{8th International Conference on Learning Representations,
  {ICLR} 2020, Addis Ababa, Ethiopia, April 26-30, 2020}. OpenReview.net.

\bibitem[{Das et~al.(2019)Das, Dhuliawala, Zaheer, and McCallum}]{das2019multi}
Rajarshi Das, Shehzaad Dhuliawala, Manzil Zaheer, and Andrew McCallum. 2019.
\newblock \href {https://openreview.net/forum?id=HkfPSh05K7} {Multi-step
  retriever-reader interaction for scalable open-domain question answering}.
\newblock In \emph{7th International Conference on Learning Representations,
  {ICLR} 2019, New Orleans, LA, USA, May 6-9, 2019}. OpenReview.net.

\bibitem[{Devlin et~al.(2019)Devlin, Chang, Lee, and
  Toutanova}]{devlin2019bert}
Jacob Devlin, Ming-Wei Chang, Kenton Lee, and Kristina Toutanova. 2019.
\newblock \href {https://doi.org/10.18653/v1/N19-1423} {{BERT}: Pre-training of
  deep bidirectional transformers for language understanding}.
\newblock In \emph{Proceedings of the 2019 Conference of the North {A}merican
  Chapter of the Association for Computational Linguistics: Human Language
  Technologies, Volume 1 (Long and Short Papers)}, pages 4171--4186,
  Minneapolis, Minnesota. Association for Computational Linguistics.

\bibitem[{Dror et~al.(2018)Dror, Baumer, Shlomov, and
  Reichart}]{dror2018hitchhiker}
Rotem Dror, Gili Baumer, Segev Shlomov, and Roi Reichart. 2018.
\newblock \href {https://doi.org/10.18653/v1/P18-1128} {The hitchhiker{'}s
  guide to testing statistical significance in natural language processing}.
\newblock In \emph{Proceedings of the 56th Annual Meeting of the Association
  for Computational Linguistics (Volume 1: Long Papers)}, pages 1383--1392,
  Melbourne, Australia. Association for Computational Linguistics.

\bibitem[{Fajcik et~al.(2021)Fajcik, Docekal, Ondrej, and
  Smrz}]{fajcik2021r2d2}
Martin Fajcik, Martin Docekal, Karel Ondrej, and Pavel Smrz. 2021.
\newblock \href {http://arxiv.org/abs/2109.03502} {R2-d2: A modular baseline
  for open-domain question answering}.
\newblock In \emph{Findings of the Association for Computational Linguistics:
  EMNLP 2021}, Punta Cana, Dominican Republic. Association for Computational
  Linguistics.

\bibitem[{Fisch et~al.(2019)Fisch, Talmor, Jia, Seo, Choi, and
  Chen}]{fisch2019mrqa}
Adam Fisch, Alon Talmor, Robin Jia, Minjoon Seo, Eunsol Choi, and Danqi Chen.
  2019.
\newblock \href {https://doi.org/10.18653/v1/D19-5801} {{MRQA} 2019 shared
  task: Evaluating generalization in reading comprehension}.
\newblock In \emph{Proceedings of the 2nd Workshop on Machine Reading for
  Question Answering}, pages 1--13, Hong Kong, China. Association for
  Computational Linguistics.

\bibitem[{Iyer et~al.(2021)Iyer, Min, Mehdad, and
  Yih}]{iyer-etal-2021-reconsider}
Srinivasan Iyer, Sewon Min, Yashar Mehdad, and Wen-tau Yih. 2021.
\newblock \href {https://www.aclweb.org/anthology/2021.naacl-main.100}
  {{RECONSIDER}: Improved re-ranking using span-focused cross-attention for
  open domain question answering}.
\newblock In \emph{Proceedings of the 2021 Conference of the North American
  Chapter of the Association for Computational Linguistics: Human Language
  Technologies}, pages 1280--1287, Online. Association for Computational
  Linguistics.

\bibitem[{Jia and Liang(2017)}]{jia2017adversarial}
Robin Jia and Percy Liang. 2017.
\newblock \href {https://doi.org/10.18653/v1/D17-1215} {Adversarial examples
  for evaluating reading comprehension systems}.
\newblock In \emph{Proceedings of the 2017 Conference on Empirical Methods in
  Natural Language Processing}, pages 2021--2031, Copenhagen, Denmark.
  Association for Computational Linguistics.

\bibitem[{Joshi et~al.(2017)Joshi, Choi, Weld, and
  Zettlemoyer}]{joshi2017triviaqa}
Mandar Joshi, Eunsol Choi, Daniel Weld, and Luke Zettlemoyer. 2017.
\newblock \href {https://doi.org/10.18653/v1/P17-1147} {{T}rivia{QA}: A large
  scale distantly supervised challenge dataset for reading comprehension}.
\newblock In \emph{Proceedings of the 55th Annual Meeting of the Association
  for Computational Linguistics (Volume 1: Long Papers)}, pages 1601--1611,
  Vancouver, Canada. Association for Computational Linguistics.

\bibitem[{Khattab et~al.(2020)Khattab, Potts, and
  Zaharia}]{khattab2020relevance}
Omar Khattab, Christopher Potts, and Matei Zaharia. 2020.
\newblock Relevance-guided supervision for openqa with colbert.
\newblock \emph{arXiv preprint arXiv:2007.00814}.

\bibitem[{Kwiatkowski et~al.(2019)Kwiatkowski, Palomaki, Redfield, Collins,
  Parikh, Alberti, Epstein, Polosukhin, Devlin, Lee, Toutanova, Jones, Kelcey,
  Chang, Dai, Uszkoreit, Le, and Petrov}]{kwiatkowski2019natural}
Tom Kwiatkowski, Jennimaria Palomaki, Olivia Redfield, Michael Collins, Ankur
  Parikh, Chris Alberti, Danielle Epstein, Illia Polosukhin, Jacob Devlin,
  Kenton Lee, Kristina Toutanova, Llion Jones, Matthew Kelcey, Ming-Wei Chang,
  Andrew~M. Dai, Jakob Uszkoreit, Quoc Le, and Slav Petrov. 2019.
\newblock \href {https://doi.org/10.1162/tacl_a_00276} {Natural questions: A
  benchmark for question answering research}.
\newblock \emph{Transactions of the Association for Computational Linguistics},
  7:452--466.

\bibitem[{Lan et~al.(2020)Lan, Chen, Goodman, Gimpel, Sharma, and
  Soricut}]{lan2019albert}
Zhenzhong Lan, Mingda Chen, Sebastian Goodman, Kevin Gimpel, Piyush Sharma, and
  Radu Soricut. 2020.
\newblock \href {https://openreview.net/forum?id=H1eA7AEtvS} {{ALBERT:} {A}
  lite {BERT} for self-supervised learning of language representations}.
\newblock In \emph{8th International Conference on Learning Representations,
  {ICLR} 2020, Addis Ababa, Ethiopia, April 26-30, 2020}. OpenReview.net.

\bibitem[{Lee et~al.(2019)Lee, Chang, and Toutanova}]{lee2019latent}
Kenton Lee, Ming-Wei Chang, and Kristina Toutanova. 2019.
\newblock \href {https://doi.org/10.18653/v1/P19-1612} {Latent retrieval for
  weakly supervised open domain question answering}.
\newblock In \emph{Proceedings of the 57th Annual Meeting of the Association
  for Computational Linguistics}, pages 6086--6096, Florence, Italy.
  Association for Computational Linguistics.

\bibitem[{Lee et~al.(2016)Lee, Salant, Kwiatkowski, Parikh, Das, and
  Berant}]{lee2016learning}
Kenton Lee, Shimi Salant, Tom Kwiatkowski, Ankur Parikh, Dipanjan Das, and
  Jonathan Berant. 2016.
\newblock Learning recurrent span representations for extractive question
  answering.
\newblock \emph{arXiv preprint arXiv:1611.01436}.

\bibitem[{Loshchilov and Hutter(2017)}]{loshchilov2018decoupled}
Ilya Loshchilov and Frank Hutter. 2017.
\newblock Fixing weight decay regularization in adam.
\newblock \emph{arXiv preprint arXiv:1711.05101}.

\bibitem[{Paszke et~al.(2019)Paszke, Gross, Massa, Lerer, Bradbury, Chanan,
  Killeen, Lin, Gimelshein, Antiga, Desmaison, K{\"{o}}pf, Yang, DeVito,
  Raison, Tejani, Chilamkurthy, Steiner, Fang, Bai, and
  Chintala}]{paszke2019pytorch}
Adam Paszke, Sam Gross, Francisco Massa, Adam Lerer, James Bradbury, Gregory
  Chanan, Trevor Killeen, Zeming Lin, Natalia Gimelshein, Luca Antiga, Alban
  Desmaison, Andreas K{\"{o}}pf, Edward Yang, Zachary DeVito, Martin Raison,
  Alykhan Tejani, Sasank Chilamkurthy, Benoit Steiner, Lu~Fang, Junjie Bai, and
  Soumith Chintala. 2019.
\newblock \href
  {https://proceedings.neurips.cc/paper/2019/hash/bdbca288fee7f92f2bfa9f7012727740-Abstract.html}
  {Pytorch: An imperative style, high-performance deep learning library}.
\newblock In \emph{Advances in Neural Information Processing Systems 32: Annual
  Conference on Neural Information Processing Systems 2019, NeurIPS 2019,
  December 8-14, 2019, Vancouver, BC, Canada}, pages 8024--8035.

\bibitem[{Pennington et~al.(2014)Pennington, Socher, and
  Manning}]{pennington2014glove}
Jeffrey Pennington, Richard Socher, and Christopher Manning. 2014.
\newblock \href {https://doi.org/10.3115/v1/D14-1162} {{G}lo{V}e: Global
  vectors for word representation}.
\newblock In \emph{Proceedings of the 2014 Conference on Empirical Methods in
  Natural Language Processing ({EMNLP})}, pages 1532--1543, Doha, Qatar.
  Association for Computational Linguistics.

\bibitem[{Rajpurkar et~al.(2018)Rajpurkar, Jia, and Liang}]{rajpurkar2018know}
Pranav Rajpurkar, Robin Jia, and Percy Liang. 2018.
\newblock \href {https://doi.org/10.18653/v1/P18-2124} {Know what you don{'}t
  know: Unanswerable questions for {SQ}u{AD}}.
\newblock In \emph{Proceedings of the 56th Annual Meeting of the Association
  for Computational Linguistics (Volume 2: Short Papers)}, pages 784--789,
  Melbourne, Australia. Association for Computational Linguistics.

\bibitem[{Rajpurkar et~al.(2016)Rajpurkar, Zhang, Lopyrev, and
  Liang}]{rajpurkar2016squad}
Pranav Rajpurkar, Jian Zhang, Konstantin Lopyrev, and Percy Liang. 2016.
\newblock \href {https://doi.org/10.18653/v1/D16-1264} {{SQ}u{AD}: 100,000+
  questions for machine comprehension of text}.
\newblock In \emph{Proceedings of the 2016 Conference on Empirical Methods in
  Natural Language Processing}, pages 2383--2392, Austin, Texas. Association
  for Computational Linguistics.

\bibitem[{Seo et~al.(2017)Seo, Kembhavi, Farhadi, and
  Hajishirzi}]{seo2016bidirectional}
Min~Joon Seo, Aniruddha Kembhavi, Ali Farhadi, and Hannaneh Hajishirzi. 2017.
\newblock \href {https://openreview.net/forum?id=HJ0UKP9ge} {Bidirectional
  attention flow for machine comprehension}.
\newblock In \emph{5th International Conference on Learning Representations,
  {ICLR} 2017, Toulon, France, April 24-26, 2017, Conference Track
  Proceedings}. OpenReview.net.

\bibitem[{Stephens(1974)}]{stephens1974edf}
Michael~A Stephens. 1974.
\newblock Edf statistics for goodness of fit and some comparisons.
\newblock \emph{Journal of the American statistical Association},
  69(347):730--737.

\bibitem[{Trischler et~al.(2017)Trischler, Wang, Yuan, Harris, Sordoni,
  Bachman, and Suleman}]{trischler2017newsqa}
Adam Trischler, Tong Wang, Xingdi Yuan, Justin Harris, Alessandro Sordoni,
  Philip Bachman, and Kaheer Suleman. 2017.
\newblock \href {https://doi.org/10.18653/v1/W17-2623} {{N}ews{QA}: A machine
  comprehension dataset}.
\newblock In \emph{Proceedings of the 2nd Workshop on Representation Learning
  for {NLP}}, pages 191--200, Vancouver, Canada. Association for Computational
  Linguistics.

\bibitem[{Vinyals et~al.(2015)Vinyals, Fortunato, and
  Jaitly}]{vinyals2015pointer}
Oriol Vinyals, Meire Fortunato, and Navdeep Jaitly. 2015.
\newblock \href
  {https://proceedings.neurips.cc/paper/2015/hash/29921001f2f04bd3baee84a12e98098f-Abstract.html}
  {Pointer networks}.
\newblock In \emph{Advances in Neural Information Processing Systems 28: Annual
  Conference on Neural Information Processing Systems 2015, December 7-12,
  2015, Montreal, Quebec, Canada}, pages 2692--2700.

\bibitem[{Wang and Jiang(2017)}]{wang2016machine}
Shuohang Wang and Jing Jiang. 2017.
\newblock \href {https://openreview.net/forum?id=B1-q5Pqxl} {Machine
  comprehension using match-lstm and answer pointer}.
\newblock In \emph{5th International Conference on Learning Representations,
  {ICLR} 2017, Toulon, France, April 24-26, 2017, Conference Track
  Proceedings}. OpenReview.net.

\bibitem[{Weissenborn et~al.(2017)Weissenborn, Wiese, and
  Seiffe}]{weissenborn2017making}
Dirk Weissenborn, Georg Wiese, and Laura Seiffe. 2017.
\newblock \href {https://doi.org/10.18653/v1/K17-1028} {Making neural {QA} as
  simple as possible but not simpler}.
\newblock In \emph{Proceedings of the 21st Conference on Computational Natural
  Language Learning ({C}o{NLL} 2017)}, pages 271--280, Vancouver, Canada.
  Association for Computational Linguistics.

\bibitem[{Wolf et~al.(2019)Wolf, Debut, Sanh, Chaumond, Delangue, Moi, Cistac,
  Rault, Louf, Funtowicz et~al.}]{wolf2019huggingface}
Thomas Wolf, Lysandre Debut, Victor Sanh, Julien Chaumond, Clement Delangue,
  Anthony Moi, Pierric Cistac, Tim Rault, R{\'e}mi Louf, Morgan Funtowicz,
  et~al. 2019.
\newblock Huggingface's transformers: State-of-the-art natural language
  processing.
\newblock \emph{ArXiv}, pages arXiv--1910.

\bibitem[{Xiong et~al.(2017)Xiong, Zhong, and Socher}]{xiong2016dynamic}
Caiming Xiong, Victor Zhong, and Richard Socher. 2017.
\newblock \href {https://openreview.net/forum?id=rJeKjwvclx} {Dynamic
  coattention networks for question answering}.
\newblock In \emph{5th International Conference on Learning Representations,
  {ICLR} 2017, Toulon, France, April 24-26, 2017, Conference Track
  Proceedings}. OpenReview.net.

\bibitem[{Xiong et~al.(2018)Xiong, Zhong, and Socher}]{xiong2017dcn+}
Caiming Xiong, Victor Zhong, and Richard Socher. 2018.
\newblock \href {https://openreview.net/forum?id=H1meywxRW} {{DCN+:} mixed
  objective and deep residual coattention for question answering}.
\newblock In \emph{6th International Conference on Learning Representations,
  {ICLR} 2018, Vancouver, BC, Canada, April 30 - May 3, 2018, Conference Track
  Proceedings}. OpenReview.net.

\bibitem[{Yang et~al.(2019)Yang, Dai, Yang, Carbonell, Salakhutdinov, and
  Le}]{yang2019xlnet}
Zhilin Yang, Zihang Dai, Yiming Yang, Jaime~G. Carbonell, Ruslan Salakhutdinov,
  and Quoc~V. Le. 2019.
\newblock \href
  {https://proceedings.neurips.cc/paper/2019/hash/dc6a7e655d7e5840e66733e9ee67cc69-Abstract.html}
  {Xlnet: Generalized autoregressive pretraining for language understanding}.
\newblock In \emph{Advances in Neural Information Processing Systems 32: Annual
  Conference on Neural Information Processing Systems 2019, NeurIPS 2019,
  December 8-14, 2019, Vancouver, BC, Canada}, pages 5754--5764.

\bibitem[{Yu et~al.(2018)Yu, Dohan, Luong, Zhao, Chen, Norouzi, and
  Le}]{yu2018qanet}
Adams~Wei Yu, David Dohan, Minh{-}Thang Luong, Rui Zhao, Kai Chen, Mohammad
  Norouzi, and Quoc~V. Le. 2018.
\newblock \href {https://openreview.net/forum?id=B14TlG-RW} {Qanet: Combining
  local convolution with global self-attention for reading comprehension}.
\newblock In \emph{6th International Conference on Learning Representations,
  {ICLR} 2018, Vancouver, BC, Canada, April 30 - May 3, 2018, Conference Track
  Proceedings}. OpenReview.net.

\end{thebibliography}
\bibliographystyle{acl_natbib}

\clearpage

\appendix
\section{Examples of Answer Span Distribution}
This section provides a deeper insight towards most probable elements of answer span PMF.\\
\\
{
\textbf{Question:} What was the first point of the Reformation?\\
\textbf{Passage:} Luther's rediscovery of "Christ and His salvation" was the first of two points that became the foundation for the Reformation. His railing against the sale of indulgences was based on it.
\\
\textbf{Ground Truth:} Christ and His salvation
}
\begin{table}[ht!]
    \begin{center}
        \begin{tabular}{cp{5cm}}
            \textbf{Confidence} & \textbf{Predictions from BERT-base}                    \\ \hline
            59.7                & Christ and His salvation" \\
            35.4                & Christ and His salvation                 \\
            2.3                 & Christ                 \\
            1.3                 & "Christ and His salvation"                                                                                            \\
            0.8                 & "Christ and His salvation                                                                            \\
            0.1                 & Christ and His salvation" was                                                                                             \\
            0.1                & "Christ                                            
            \end{tabular}
    \captionof{figure}{\label{fig:indep_error2a}Example of answer span distribution from model trained via \textit{independent} objective.}
    \end{center}
\end{table}
\begin{table}[ht!]
    \begin{center}
        \begin{tabular}{cp{5cm}}
            \textbf{Confidence} & \textbf{\thead{Predictions from \\ BERT-base-\textit{compound}} }                   \\ \hline
            71.8                & Christ and His salvation \\
            10.9                & "Christ and His salvation"                 \\
            4.7                & Christ and His salvation"                 \\
            4.6                 & Luther's rediscovery of "Christ and His salvation                                         \\
            3.1                 & "Christ and His salvation                                                                            \\
            1.2                 & Luther's rediscovery of "Christ and His salvation"                                  \\
            0.8                 & Luther's rediscovery                               
            \end{tabular}
    \captionof{figure}{\label{fig:indep_error2b}Example of answer span distribution from model trained via \textit{compound} objective.}
    \end{center}
\end{table}

\vfill\eject
{
\noindent 
\textbf{Question:} How many species of bird and mammals are there in the Amazon region?\\
\textbf{Passage:} The region is home to about 2.5 million insect species, tens of thousands of plants, and some 2,000 birds and mammals. To date, at least 40,000 plant species, 2,200 fishes, 1,294 birds, 427 mammals, 428 amphibians, and 378 reptiles have been scientifically classified in the region. One in five of all the bird species in the world live in the rainforests of the Amazon, and one in five of the fish species live in Amazonian rivers and streams. Scientists have described between 96,660 and 128,843 invertebrate species in Brazil alone.
\\
\textbf{Ground Truth:} 2,000
}
\begin{table}[ht!]
    \begin{center}
        \begin{tabular}{cp{5cm}}
            \textbf{Confidence} & \textbf{Predictions from BERT-base}                    \\ \hline
            37.0                & 2,000 birds and mammals. To date, at least 40,000 plant species, 2,200 fishes, 1,294 birds, 427 \\
            34.6                & 427                 \\
            27.7                & 2,000                 \\
            0.2                 & 1,294 birds, 427                                                                                            \\
            0.2                 & 427 mammals                                                                            \\
            0.1                 & 2,000 birds                                                                                             \\
            0.1                 & 2,000 birds and mammals                                            
            \end{tabular}
    \captionof{figure}{\label{fig:indep_error3a}Example of answer span distribution from model trained via \textit{independent} objective.}
    \end{center}
\end{table}
\begin{table}[ht!]
    \begin{center}
        \begin{tabular}{cp{5cm}}
            \textbf{Confidence} & \textbf{\thead{Predictions from \\ BERT-base-\textit{compound}} }                   \\ \hline
            71.7                & 427 \\
            21.5                & 2,000                 \\
            5.1                & 2,000 birds and mammals. To date, at least 40,000 plant species, 2,200 fishes, 1,294 birds, 427                 \\
            0.8                 & some 2,000                                                                                              \\
            0.2                 & 427 mammals                                                                            \\
            0.1                 & 1,294 birds, 427                                                                                              \\
            0.1                 & 2,000 birds and mammals. To date, at least 40,000 plant species, 2,200 fishes, 1,294                                            
            \end{tabular}
    \captionof{figure}{\label{fig:indep_error3b}Example of answer span distribution from model trained via \textit{compound} objective.}
    \end{center}
\end{table}

\vfill\eject
\pagebreak

\section{Conditional Objective}
\label{appendix:conditional_obj}
Some of the recent LRMs assume conditional factorization of span's PMF. For comparison with our joint objective, we reimplemented the conditional objective used in ALBERT \cite{lan2019albert}.

First, the probabilities $\boldsymbol{P}(a_s)$ for the start position are computed in the same manner as for the independent objective -- by applying a linear transformation layer on top of representations $\boldsymbol{H} \in \mathbb{R}^{d \times L}$ from the last layer of the LRM, where $d$ is the model dimension and $L$ denotes the input sequence length.
\begin{equation}
    \boldsymbol{P}(a_s)\propto \exp{(\boldsymbol{w}_s^\top\boldsymbol{H} + \boldsymbol{b_s})}
\end{equation}
During the validation, top $k$ ($k=10$ in our experiments) start positions are selected from these probabilities, while in the training phase, we apply teacher forcing by only selecting the correct start position. Representation of i-th start position $\boldsymbol{h}_i$ from the last layer of the LRM corresponding to the selected position is then concatenated with representations corresponding to all the other positions $k={0..L}$ into matrix $\boldsymbol{C}$. 

\begin{equation}
\boldsymbol{C}= 
\begin{bmatrix}
    \text{---} & [\boldsymbol{h_0};\boldsymbol{h_i}] & \text{---} \\
    \text{---} & [\boldsymbol{h_1};\boldsymbol{h_i}] & \text{---} \\
    \multicolumn{3}{c}{$\vdots$}   \\
    \text{---} & [\boldsymbol{h_n};\boldsymbol{h_i}] & \text{---} 
\end{bmatrix}
\end{equation}

Subsequently, a layer with $\tanh$ activation is applied on this matrix $\boldsymbol{C}$, followed by a linear transformation to obtain the end probabilities:
\begin{equation}
    \boldsymbol{P}(a_e|a_s=i) \propto \exp{( \boldsymbol{w}_c^\top  \tanh{(\boldsymbol{W}\boldsymbol{C} + \boldsymbol{b'}) + \boldsymbol{b}})}
\end{equation}

For each start position we again select top k end positions, to obtain $k^2$-best list of answer spans. In contrast to the official ALBERT implementation, we omitted a layer normalization after \textit{tanh} layer.
\begin{table*}[!ht]
    \centering
    \begin{tabular}{|c|c|c|c|c|}
\hline
\textbf{Model}&   & \textbf{I}                   & \textbf{J}                      & \textbf{I+J}    \\ \hline
\multirow{2}{*}{BIDAF}               
              & LF & 66.16/76.19                   & 58.24/67.42                     & 66.96/75.90   \\ \cline{2-5} 
              & SF & 66.20/76.21                   & -                               & 66.99/75.90   \\ \hline\hline
\multirow{2}{*}{BERT} 
              & LF & 81.31/88.65                   & 81.33/88.13                     & 81.83/88.52   \\ \cline{2-5} 
              & SF & 81.38/88.68                   & 81.23/87.97                     & 81.65/88.36   \\ \hline\hline
\multirow{2}{*}{ALBERT}               
              & LF & 88.55/94.63                   & 88.84/94.64                     & 89.02/94.77   \\ \cline{2-5} 
              & SF & 88.53/94.00                   & 88.28/94.10                     & 88.68/94.49   \\ \hline
\end{tabular}
    \caption{\label{tab:sff_results} SQuADv1.1 EM/F1 results with length filtering (LF) and LF + surface form filtering (SF).}
\end{table*}
\section{Addressing the Complexity}
\label{sec:adressing_the_complexity}
One may ask what complexity joint modelling objectives come with independently of the underlying architecture. Given that $L$ is the length of the input's passage and $d$ is the model dimension, the independent objective contains only linear transformation and is in $\BigO{d L}$ for time and memory, assuming the multiplication and addition are constant operations. For the rest of this analysis, we will denote both time and memory complexities as just complexity, as they are the same for the analyzed cases.

The conditional objective increases the complexity for both only constantly, having an extra feed-forward network for end token representations. However, one may experience a significant computational slowdown, because of the beam search. Having a beam size $k$  and a minibatch size $b$, the end probabilities cannot be computed in parallel with start probabilities, and have to be computed for the $kb$ cases.

For the direct joint probability modelling, the complexity largely depends on the similarity function. The easiest case is $f_{dot}$, where in theory the complexity rises to $\BigO{d L^2}$, but in practice the dot product is well optimized and has a barely noticeable impact on the speed or memory.

For the $f_{add}$ the complexity is given by the linear projection $\boldsymbol{H_* w_*}$ being in $\BigO{dL}$ and outer summation of two vectors $\boldsymbol{H_s w_1} \oplus \boldsymbol{H_e w_2}$, which is in $\BigO{L^2}$, where $\boldsymbol{w} = [\boldsymbol{w_1}, \boldsymbol{w_2}]$ and $\boldsymbol{H_*} \in \mathbb{R}^{n \times d}$ are the start/end representation matrices. Therefore the complexity is $\BigO{dL + L^2}$. We observed that in practice this approach is not very different from $f_{dot}$, probably due to $d$ being close to $L$.

Next, a weighted product $f_{wdot}$ can be efficiently implemented as $\boldsymbol{H_s}(\boldsymbol{w} \circ \boldsymbol{H_e})$, where $\boldsymbol{w}$ is broadcasted over every end representation in $\boldsymbol{H_e}$. In this case, the complexity stays the same as for $f_{dot}$.

To demonstrate that in practice the speed and memory requirements between independent and joint approach are comparable, one BERT epoch on SQuADv1.1 took about 47 minutes and 4.2GB of memory with the same batch size 2 on 12GB 2080Ti GPU with both objective variants. We observed the same requirements for all direct joint probability modelling methods mentioned so far.

Finally, the most complex approach is clearly $f_{MLP}$. While an a theoretical time and memory complexity of an efficient implementation\footnote{The linear transformation $d\times 2d$ can be applied to each start or end vector separately, and only then the start/end vectors have to be outer-summed.} is in $\BigO{d^2L + dL^2}$, the complexity of this approach can be improved by pruning down the number of possible spans (and the probability space). Assuming the maximum length of the span is $k \ll L$, one can reduce the complexity to $\BigO{d^2L + dLk}$ (an approach adopted in \citet{lee2019latent}). 
To illustrate this complexity, BERT model with the full probability space on SQuADv1.1 with batch size 2 took 76 minutes per epoch while allocating 8.2GB of GPU memory (we were unable to fit larger batch size to 12GB GPU).

\section{Hyperparameters}
\label{section:hyperparameters}
The exact hyperparameters used in this work are documented in our code. We note that for BERT and ALBERT, we simply followed the hyperparameters proposed by the authors for SQuADv1.1. In case of LRM models, each input context is split into windows as proposed by \citet{devlin2019bert}. Each input sequence has maximum length 384, questions are truncated to 64 tokens and context is split with overlap stride 128. For SQuADv2.0, we follow the BERT's approach for computing the no-answer logit in test-time. Having the set of k windows $W_e$ for each example $e$, we compute the null-score $ns_w=\textrm{logit}P(a_s=0) + \textrm{logit}P(a_e=0)$ for each window $w \in W_e$. For joint and compound objectives $ns_w=\textrm{logit} P(a_s=0, a_e=0)$. Defining that for each window $w$ the best non-null answer logit is $a_w$, the no-answer logit is then given by the difference of lowest null-score and best-answer score $\Gamma = \min_{w\in W_e}(ns_w) - \max_{w\in W_e}(a_w)$ among all windows of example $e$. The threshold for $\Gamma$ is determined on the validation data via official script.

\section{Marginalizing Over the Same String Forms}
\label{appendix:sff}
To alleviate the error type (3) from section \ref{section:analysis}, we experimented with marginalizing over probabilities of top-100 answers (so-called surface form filtering). This is done via summing the probabilities into the most probable string occurrence, and setting the probability of the rest to 0. The results for all trained models averaged over 10 checkpoints are presented in Table \ref{tab:sff_results}. Note this approach sometimes hurts performance, especially in  the case of joint probability approaches, where this error type happens very rarely.

\end{document}